\documentclass[11pt]{article}

\usepackage[preprint]{acl}

\usepackage{times}
\usepackage{latexsym}

\usepackage[T1]{fontenc}

\usepackage[utf8]{inputenc}

\usepackage{microtype}
\usepackage{inconsolata}
\usepackage{graphicx}
\usepackage[export]{adjustbox} 
\usepackage{stfloats}

\usepackage{enumitem}

\title{Pterodactyl: Pushing the Limits of Efficient Automated Translation Evaluation}
\author{
\textbf{Evelyn Y. Garland\textsuperscript{1, 2}} and
\textbf{Carola F. Berger\textsuperscript{1, 3}}
\\
\textsuperscript{1}1020 Group,
\textsuperscript{2}Acta Language Services, LLC,
\textsuperscript{3}CFB Scientific Translations LLC
\\
\small{
 \textbf{Correspondence:} \href{mailto:project1020group@gmail.com}{project1020group@gmail.com}
  }
}

\begin{document}
\maketitle

\begin{abstract}
In this paper, we present 1020 Group's submission to the WMT26 shared task on automated quality evaluation. Our system, Pterodactyl, was engineered specifically for subtasks 2 and 3, that is, segment-level quality score prediction and the detection of error-free segments. Pterodactyl is a highly customizable, ultra-low-cost system optimized for binary classification in translation triage and data curation. It is based on an LLM-as-a-judge approach, with prompt engineering and model selection guided by ROC analysis. We also developed several variants of Pterodactyl that were specifically adapted for subtasks 3. 
\end{abstract}

\section{Introduction}

In recent WMT competitions, reference-free quality estimation (QE) systems based on an LLM-as-a-judge approach have gained increasing traction for automated translation quality evaluation~\cite{lavie-etal-2025-findings,freitag-etal-2024-llms, zerva-etal-2024-findings}. However, the development and deployment of such systems can be complex and costly. Here, we develop an efficient, production-oriented strategy for pushing the cost-performance frontier of automated translation evaluation: achieving better task-specific performance with faster and cheaper models and prompt configurations.

Below, we present 1020 Group's submission to the WMT26 shared task on automated quality evaluation. Our system, Pterodactyl, and its variants are intended as highly customizable, ultra-low-cost, and robust options for performing segment-level score prediction (WMT26 automated quality evaluation subtask 2) and detecting error-free segments (WMT26 automated quality evaluation subtask 3). The system is highly customizable thanks to an efficient, streamlined engineering process that can be implemented to select and build fit-for-purpose QE systems for different application scenarios. When possible, we avoid separately optimizing every component when a shared screening criterion might suffice. Pterodactyl and its variants are also ultra-low-cost, in terms of both time and tokens, because our approach allows us to minimize cost and latency when two or more options show similar performance on test data. Finally, our approach allows engineering a system that remains relatively stable in different testing scenarios and under variations in the gold standard caused by inter-annotator disagreement.

\section{The Pterodactyl Premise}

Our approach is based on the LLM-as-a-judge paradigm, whereby our prompt engineering methodology and the selection of the most suitable models were guided by Receiver Operating Characteristic (ROC) analysis~\cite{FAWCETT2006861}, supplemented by additional considerations. Our goal was to engineer a robust system that is  simple to implement and adapt as well as sufficiently low-cost to be used in large production runs. 

\subsection{Basic Considerations}\label{sec:considerations}

To develop our system, we started with commercially available general-purpose models and performed prompt engineering based on the considerations detailed below. 

Our starting point was the basic premise that a QE system that is able to separate good and bad translations well is a promising candidate for WMT26 subtasks 2 and 3. We have shown in~\cite{garlandberger2026} that ROC analysis provides an efficient screening tool to find such candidates among a pool of systems. Prompt and model selection for subtasks 2 and 3 can therefore be partially unified with this ROC screening method.

We tested the following additional considerations during our prompt engineering iterations for both subtasks 2 and 3:
\begin{itemize}[nosep]
\item Whether newer (larger) frontier models are better for translation quality estimation
\item Whether more reasoning improves estimation performance
\end{itemize}

\subsection{Further Considerations for Subtask 3}\label{sec:MCCselection}

For subtask 3, further considerations come into play. Here, translations are to be triaged into error-containing and error-free segments, with a human gold standard obtained via contrastive error span annotation (cESA)~\cite{zouhar2026contrastiveesahumanevaluation}. Submissions for subtask 3 are evaluated and ranked with reference to this cESA gold standard, as measured by the Matthews Correlation Coefficient (MCC). Since cESA is a new human annotation method for which no extensive development data is available, our strategy was to optimize our system for robustness with respect to the exact score threshold that separates the error-free class from the error-containing one.

\begin{figure*}[b!]
\centering
  \includegraphics[width=0.48\linewidth, valign=c]{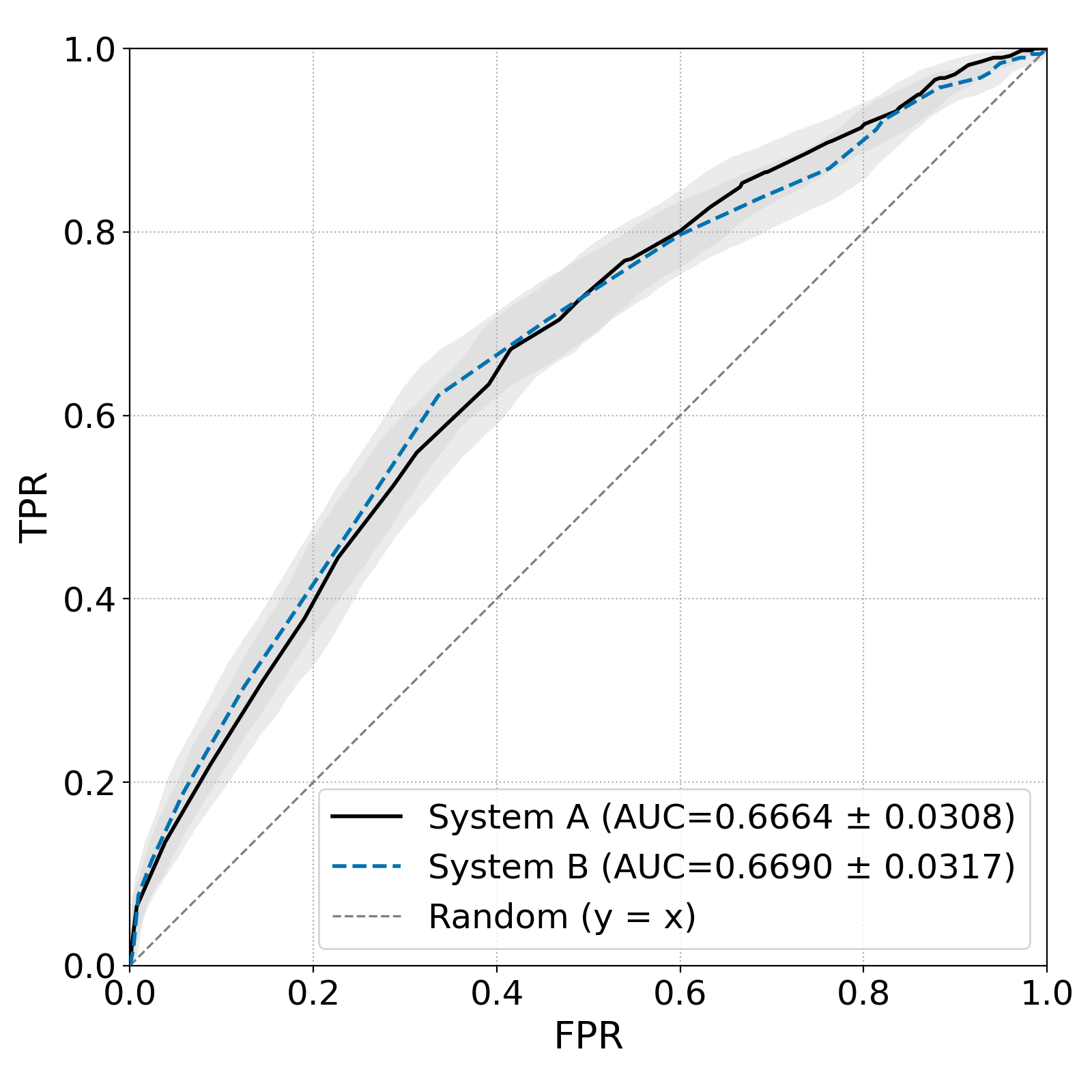} \hfill
  \includegraphics[width=0.48\linewidth, valign=c]{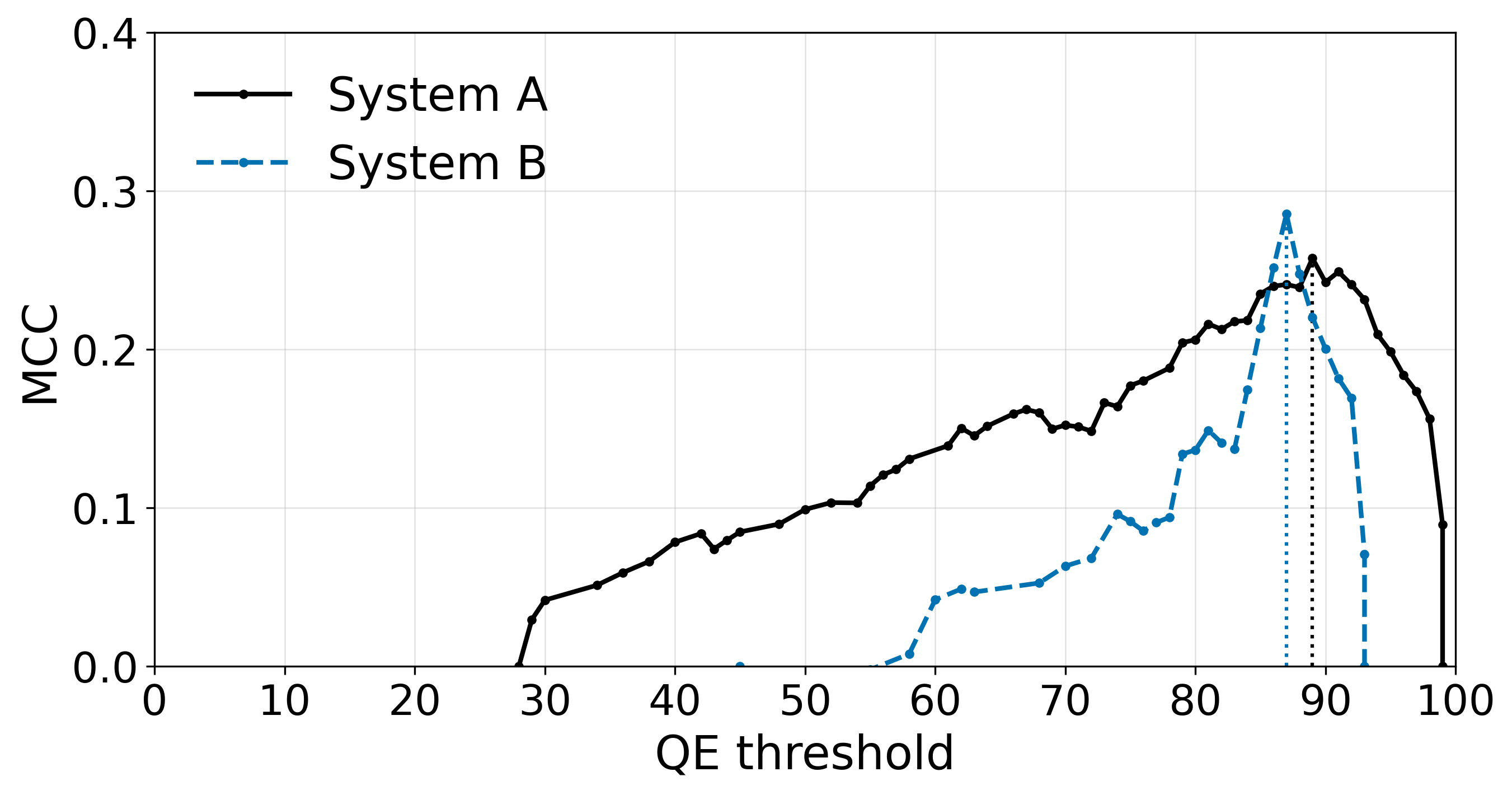}
  \caption {ROC curves (left) and MCC versus error-classifying threshold curves (right) for two example systems.}
  \label{fig:MCCplateau}
\end{figure*}

What do we mean by robustness? Past WMT tasks used standard error span annotation (ESA)~\cite{kocmi-etal-2024-error}, with slightly different guidelines and anchor specifications (see, e.g. ~\cite{lavie-etal-2025-findings} and references therein). It is therefore a priori unknowable if and how the WMT26 gold standard differs from prior years' ESA gold standards. Thus, an error-classifying threshold that may be optimal for our system for WMT25 data may not be entirely optimal for WMT26. As a consequence, we did not optimize our prompts to obtain maximum MCC when computed on earlier WMT development data, but rather to obtain a robustness plateau, that is, a range within which small variations in the error-classifying threshold lead to relatively minor changes in the MCC value. This can be observed in Figure~\ref{fig:MCCplateau}. 

Figure~\ref{fig:MCCplateau} shows two different systems with similar ROC performance, but quite different MCC versus threshold behavior. Example Systems A and B have comparable ROC curves and ROC-AUC values, but their MCC versus QE threshold behavior differs. While System A has a slightly lower best MCC value compared to System B, the peak of System A's MCC distribution is decidedly broader than that of System B.

\section{Our System Engineering Approach} 

\subsection{Sample Data and Baselines} 

For the prompt engineering iterations described in the next subsection, we used data from past Conferences on Machine Translation (WMT) that were available in a public GitHub repository~{\tt https://github.com/google\-research/mt-metrics-eval}, licensed under Apache v2.0. Specifically, we used the English into Chinese data with ESA annotations from WMT25~\cite{lavie-etal-2025-findings}, and the English into German data from WMT23 in the same repository that were supplemented with ESA annotations by~\cite{kocmi-etal-2024-error,zouharkocmi2024esaai}. These ESA annotations for English into German WMT23  data are available at~{\tt https://github.com/wmt-conference/\-error\-spanannotation}. For our prompt iterations, we split each of these datasets into a development set containing approximately 70\% of the en$\rightarrow$zh and en$\rightarrow$de datasets, respectively, and a test set containing the remaining 30\%. 

Our main gold standard was defined in a manner similar to WMT26 subtask 3: An error-free segment is defined as having an ESA score $\geq$ 85 and no error span annotation. Where two sets of human annotations were available, we took the average of the ESA scores $\geq$ 85, and either no error span annotation in annotation set 1 or no error span annotation in set 2. It turned out that our relative screening method via ROC analysis is relatively robust to changes in the definition of the gold standard.

And finally, to study an example of an ultra-low-resource language for subtask 3, we used the English into Northern S\'{a}mi recipe~{\tt wmt26-eng-sme} from the WMT26 General MT task datasets. From this recipe, we created a somewhat artificial development set as follows: We randomly selected 2000 longer segments from the recipe, then randomly selected 1000 segments from these as an error-free gold standard, and modified the other 1000 segments with random punctuation errors, omissions and additions  of random words in Northern S\'{a}mi, as well as random source text insertions to obtain 1000 error-containing gold standard segments. Of course, this is nowhere near equivalent to a real development dataset, but given the fact that none of the authors speaks Northern S\'{a}mi, this is the closest available approximation.

As our baselines, we ran CometKiwi22~\cite{rei-etal-2022-cometkiwi} and xCOMET-XL~\cite{guerreiro2023xcomettransparentmachinetranslation} on these development and test data. 

For testing and developing prompts for English into Northern S\'{a}mi, we also utilized the open-source LLM NorMistral-11B~\cite{samuel2025fluentalignmentdisfluentjudges,samuel-etal-2025-small}, with prompts in the Norwegian language instead of English.

\subsection{Iterative Prompt Engineering}

Our prompt engineering approach was iterative, discarding prompts that are no better than our baselines, while attempting to improve the promising ones that exhibited the best ROC-AUC performance. Once this iteration converged to a few prompts with similar ROC performance, we selected those that showed a robust MCC performance as detailed in Section~\ref{sec:MCCselection}. As our final system and system variants, we chose the ones that exhibited the minimum cost and latency. This highly efficient approach can be automated and easily adapted to other scenarios outside the WMT realm.

We experimented with a variety of prompts. On the basis on linguistic and other technical considerations, we identified six distinct main strategies. After performing ROC analysis on the results, we discarded those that were clearly inferior. On the prompts that seemed promising, we iterated a number of times, incorporating various refinements. Our findings are reported in the next section.

\section{Submitted Systems}

Using the aforementioned iterative prompt engineering approach, we were able to answer some of the questions in Section~\ref{sec:considerations} and converge on a final system.  The one question we were unable to answer is whether the newest models, which were presumably trained on the largest and most recent training data, are able to handle ultra-low-resource languages better than slightly older models, given the same prompt. This led us to submit two system variants to subtask 2 (see Table~\ref{systemtable}), the primary system using the same cost-efficient general-purpose model for all languages, and a variant using one of the newest available models for the three language pairs English into Ladin, Ligurian, and Northern S\'{a}mi. The official results of subtask 2 will tell which option is better.

As a side note, while we used NorMistral-11B for testing and development, we did not end up incorporating it into our system framework. This is because we did not perform a separate prompt engineering optimization for Northern S\'{a}mi, in part because Norwegian seems to be a better prompting language for NorMistral-11B than English, and none of the authors speaks Norwegian. With the same prompt as for all the other languages, the commercial general purpose models we studied seem to perform better on our artificial mock dataset for Northern S\'{a}mi, but this may well be because the prompt was optimized for these models, not for NorMistral-11B, or perhaps because our artificial mock dataset is constructed somewhat arbitrarily, or a combination of both.

\begin{figure}[htb!]
\centering
 \includegraphics[width=0.9\columnwidth]{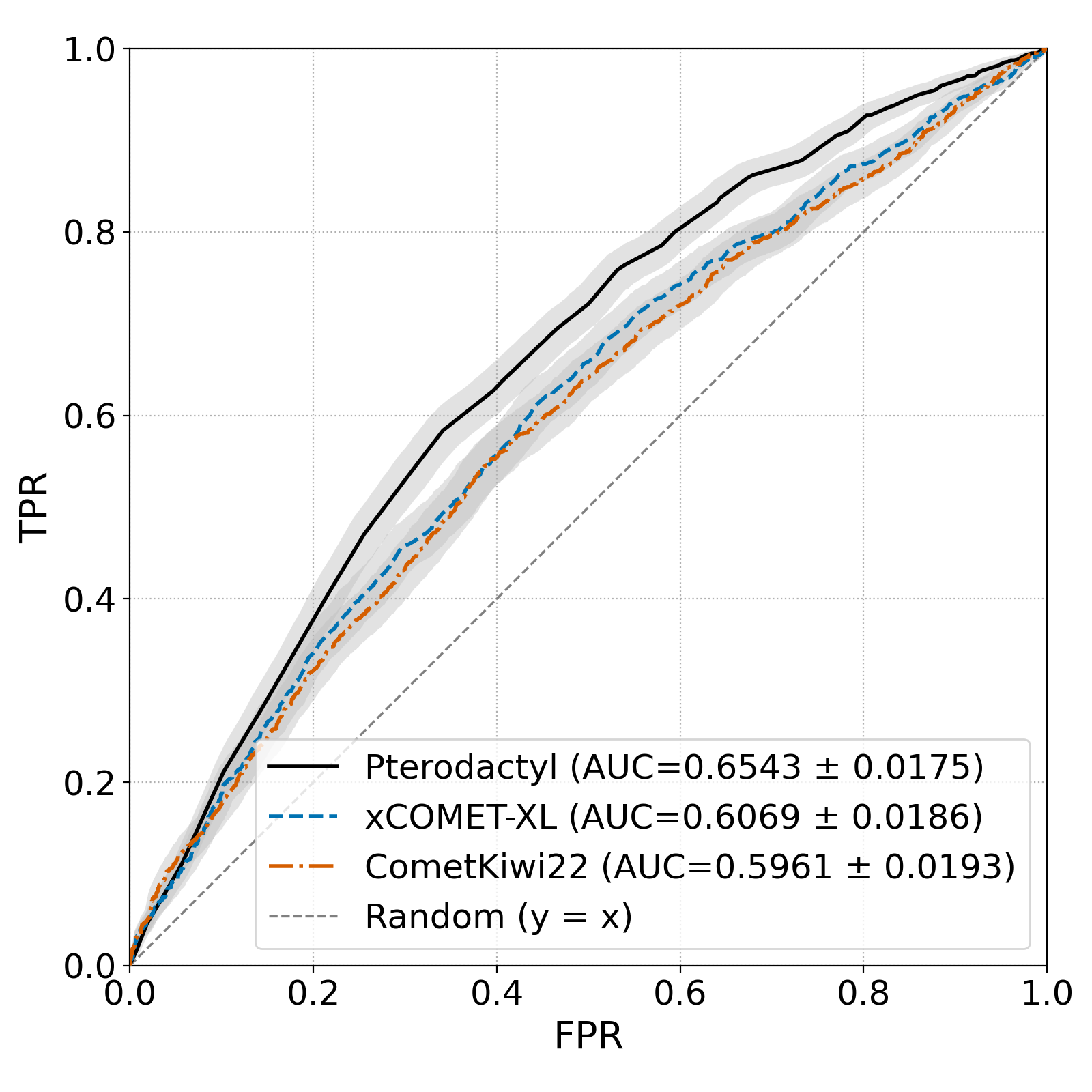} 
  \caption{ROC curve of the chosen prompt and model combination compared to the baselines CometKiwi22 and xCOMET-XL on the WMT25 en$\rightarrow$zh dataset.}
  \label{fig:Ptero}
\end{figure}

\begin{figure}[htb!]
\centering
 \includegraphics[width=0.9\columnwidth]{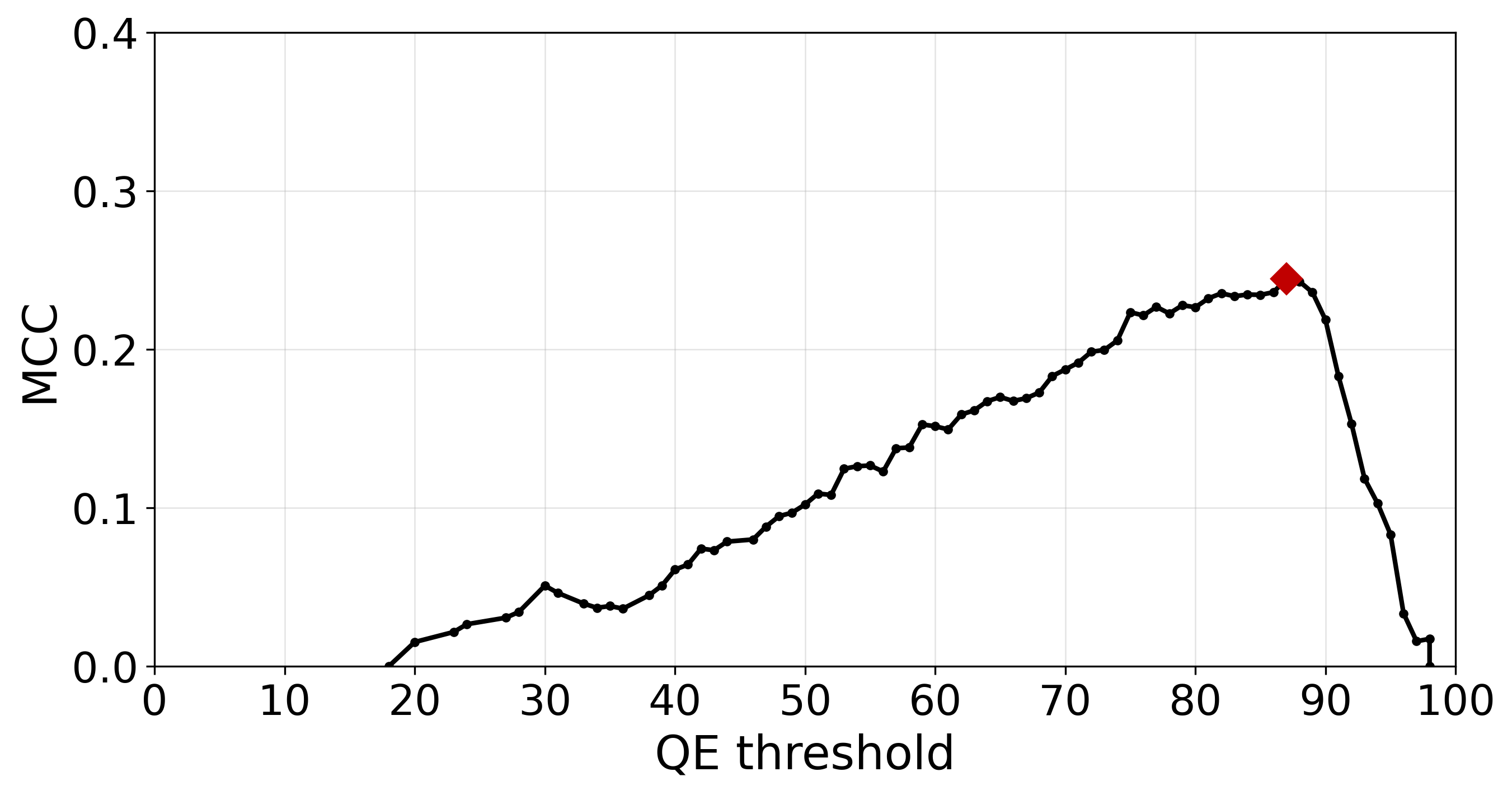} 
  \caption{MCC versus error-classifying threshold curve for the selected system on the WMT25 en$\rightarrow$zh dataset. The red diamond indicates the maximum MCC value.}
  \label{fig:Ptero2}
\end{figure}

The final selected prompt instructs the LLM to evaluate the translation quality for each segment individually, without requiring the model to identify specific translation errors or to justify its score. Figure~\ref{fig:Ptero} shows the selected system's ROC curve compared to our baselines, CometKiwi22 and xCOMET-XL, on the WMT25 English into Chinese data. Figure~\ref{fig:Ptero} also lists the corresponding ROC-AUC values with their 95\% confidence intervals. Figure~\ref{fig:Ptero2} shows the selected system's MCC versus threshold behavior. 

We ran our scoring in batches of up to 20 segments produced by the same MT system for the same language pair, because this batching procedure leads to lower token usage and cost. 

For subtask 3, we submitted two system variants that are derived from Pterodactyl and specifically optimized for subtask 3, employing a threshold-setting strategy based on patterns observed in the WMT25 data.

Table~\ref{systemtable} gives an overview of our primary system and its variant that were submitted to subtasks 2 and 3, along with the computational cost involved in scoring the official WMT26 test data. Table~\ref{systemtable3} lists the variants we submitted specifically to subtask 3.

\begin{table*}[bht!]
  \centering
  \begin{tabular}{llll}
    \hline
  & &  \textbf{Pterodactyl} & \textbf{Austriadactylus} \\
\hline
  \multicolumn{2}{l}{  \textbf{Description}} & Primary system,  & Variant,  \\
 &  & optimized for subtasks 2 \& 3 & optimized for subtasks 2 \& 3 \\ \hline
  \multicolumn{2}{l}{   \textbf{Foundational LLM} } & OpenAI gpt-5.4 & OpenAI gpt-5.4  \\ \hline
  \multicolumn{2}{l}{  \textbf{Reasoning effort} } &  None & None  \\ \hline
\multicolumn{2}{l}{     \textbf{Language-specific } } & None & OpenAI gpt-5.6-terra,  \\
\multicolumn{2}{l}{     \textbf{exceptions} } & & low reasoning, for \\
&    & &  Ladin, Ligurian, Northern S\'{a}mi  \\
    \hline
    \textbf{Computational cost} & \textbf{Total input} & 68.2 & 68.4 \\
    \textbf{[million tokens]} & \textbf{Total output} & 10.7 & 11.5  \\ \hline
    \end{tabular}
\caption{Systems submitted by 1020 Group to WMT26 subtasks 2 and 3, along with the computational cost involved in scoring the official WMT26 test data.}
  \label{systemtable}
\end{table*}

\begin{table*}[bht!]
  \centering
  \begin{tabular}{lll}
    \hline
 &  \textbf{Wukongopterus} & \textbf{Austriadraco} \\
\hline
    \textbf{Description} & Variant,  & Variant,  differently \\
    & optimized for subtask 3 & optimized for subtask 3 \\ \hline
    \textbf{Foundational LLM} & OpenAI gpt-5.4 & OpenAI gpt-5.4  \\ \hline
    \textbf{Reasoning effort} &  None & None  \\ \hline
    \textbf{Language-specific } & None & None  \\
    \textbf{exceptions}  & &  \\    \hline
    \end{tabular}
\caption{Systems submitted by 1020 Group to WMT26 subtask 3. No additional LLM-computational cost is involved, since the scores are taken from the main Pterodactyl scoring run and subsequently converted into a binary classification for subtask 3 via Python.}
  \label{systemtable3}
\end{table*}

\section{Conclusions and Outlook}

Pterodactyl's central design principle was deliberate minimalism: use no more model or engineering complexity than the task requires. Our submission to subtasks 2 and 3 of the WMT26 shared task on automated quality evaluation is based on an LLM-as-a-judge approach. Our prompt engineering and model selection process uses ROC analysis as a screening tool, making the development highly adaptable, robust, and low-cost. It remains to be seen, however, whether this minimal approach is an effective alternative to fine-tuning or building from scratch. The shared-task results will show whether Pterodactyl merely costs less---or whether it offers a genuinely competitive alternative to more resource-intensive system development.

The astute reader will of course observe that we have committed the cardinal sin of extrapolating from not necessarily representative development data into the great unknown of WMT26, which includes new language pairs and a novel human annotation scheme. But this leap of faith is not entirely unfounded, based on high-level patterns we observed across languages and domains from the WMT25 and WMT23 data, supplemented by our ``artificial'' Northern S\'{a}mi data. In an actual production environment, such an extrapolation is of course unnecessary, and the system can easily be adapted to real in-domain and in-language data for production scenarios, as described above.

\section{Limitations}

As already mentioned, the exact human gold standard for WMT26 is unknowable, because in prior WMT tasks, standard ESA was employed, whereas this year's task utilizes contrastive ESA for the first time. However, this unknown factor is the same for all participants, and it remains to be seen what, if any, effects the new human annotation scheme has on the results. Another unknown is how this year's test dataset, including the addition of new language pairs without human-annotated data, differs from those of previous years. Again, all participants are in the same boat here. Finally, our system was developed with rather limited resources, and we tested only on two language pairs with ESA annotations. Further studies are needed to show whether our one-size-fits-all system really does fit all.

\section*{Acknowledgments}

If we score higher, it is because we have the privilege of choosing which giants' shoulders we stand on. We are deeply grateful to those who made that choice possible. 
We also thank Prof. Alon Lavie for encouraging us, and the WMT organizers for enabling us to participate and for openly providing past years? data for development purposes. 

\bibliography{Pterodactyl_1020Group}

\end{document}